\documentclass[letterpaper, 10 pt, conference]{ieeeconf}  

\IEEEoverridecommandlockouts                              

\usepackage[colorlinks=false]{hyperref}
\usepackage[pdftex]{graphicx}
\usepackage[space]{cite}
\usepackage{booktabs}
\usepackage{amsmath,amssymb}
\usepackage{xcolor}
\usepackage{multirow}
\usepackage{color,soul}

\hypersetup{
	hidelinks,
	colorlinks=false,
	linkcolor=black,
	citecolor=black
}

\title{\LARGE \bf An Efficient Plane Extraction Approach for Bundle Adjustment on LiDAR Point clouds}

\author{Zheng Liu$^{1}$ and Fu Zhang$^{1}$
\thanks{$^{1}$Zheng Liu and Fu Zhang are with the Department of Mechanical Engineering, The University of Hong Kong, Hong Kong.
        {\tt\small u3007335@connect.hku.hk}, {\tt\small fuzhang@hku.hk}}
}

\begin{document}
\maketitle
\thispagestyle{empty}
\pagestyle{empty}

\begin{abstract}


Bundle adjustment (BA) on LiDAR point clouds has been extensively investigated in recent years due to its ability to optimize multiple poses together, resulting in high accuracy and global consistency for point cloud. However, the accuracy and speed of LiDAR bundle adjustment depend on the quality of plane extraction, which provides point association for LiDAR BA. In this study, we propose a novel and efficient voxel-based approach for plane extraction that is specially designed to provide point association for LiDAR bundle adjustment. To begin, we partition the space into multiple voxels of a fixed size and then split these root voxels based on whether the points are on the same plane, using an octree structure. We also design a novel plane determination method based on principle component analysis (PCA), which segments the points into four even quarters and compare their minimum eigenvalues with that of the initial point cloud. Finally, we adopt a plane merging method to prevent too many small planes from being in a single voxel, which can increase the optimization time required for BA. Our experimental results on HILTI demonstrate that our approach achieves the best precision and least time cost compared to other plane extraction methods.

\end{abstract}

\section{Introduction}

In recent years, light detection and ranging (LiDAR) technology has become increasingly popular in the robotics community \cite{urmson2008autonomous} due to its ability to provide direct, dense, active and accurate depth measurements of environments. As a result, LiDAR plays an significant role in many mobile robotic applications, such as self-driving vehicles \cite{hecht2018lidar} and autonomous UAVs \cite{kong2021avoiding, gao2019flying}. The high cost of traditional mechanical LiDARs has led to increased  research interests in solid-state LiDAR, which is lightweight, mass-produced and low cost \cite{liu2021low, wang2020mems}, making it a more accessible option for researchers. 


{The mainstream point cloud registration approaches} \cite{besl1992method, segal2009generalized, biber2003normal} {can only perform pair-wise registration incrementally, which can result in cumulative drift.} To address this issue, the bundle adjustment (BA) technique, which is widely used in visual SLAM, has been investigated to eliminate the cumulative drift, ensuring global consistency of the LiDAR point cloud in many works \cite{liu2022efficient, zhou2020efficient, ferrer2019eigen, huang2021bundle, liu2021balm}. The advantage of BA is capable of optimizing multiple frame poses and the geometry of landmarks together. Moreover, most of the existing LiDAR BA formulations are based on the plane model to build the cost function. Hence, detecting planes quickly and accurately is important for LiDAR BA to provide enough planes as point association, and the quality of detected plane will also affect the precision and efficiency of LiDAR BA.

Plane extraction is a classic and important problem in point cloud applications, such as SLAM \cite{behley2018efficient, yuan2022efficient}, 3D reconstruction \cite{vizzo2021poisson}, and scene recognition \cite{el2019rgb}. Compared to raw point samples, planes can represent the geometric information of multiple points with a simple parametric model, reducing computational time and the noise. Consequently, plane extraction from point clouds has been extensively investigated, and most methods can be classified into three main categories: Hough transform (HT), random sample consensus (RANSAC) and region growing (RG). However, these traditional methods are not directly applicable to LiDAR BA due to the efficiency and quality.

For this issue, we present an efficient plane extraction approach for LiDAR BA in this work, building on the adaptive voxelization method in BALM \cite{liu2021balm}. We begin by voxelizing the space using a default size (eg., $1m$ or $2m$), and then iterate through each root voxel to determine whether its points lie on the same plane. If all points lie on one plane, current voxel is kept and considered as a plane; If not, the current voxel is partitioned into eight octants, and the process is repeated for each non-empty octant until the points are on the same plane or meet the termination condition (e.g., minimal point number or octant size). To more accurately determine whether the points are on the same plane, we design a novel plane determination method based on principle component analysis (PCA). This method segments the points into four even quarters and compares their minimum eigenvalues with that of the initial point cloud. After the partition of octree, we merge the small planes in a root voxel into larger ones to reduce the time cost of LiDAR BA optimization. 

The reminder of this work is organized as follows. Section \ref{related work} introduces the related papers of LiDAR bundle adjustment and plane extraction. In Section \ref{methodology}, we present our methodology. The experimental results are presented in Section \ref{experiment}. Finally, in Section \ref{conslusion}, we conclude this article and propose directions for future work.

\section{Related work} \label{related work}

Plane extraction has been extensively investigated and applied in numerous point cloud applications. However, different usages require different planar quality and extraction speed. In this paper, we primarily focus on the planes requirement of LiDAR bundle adjustment.

\subsection{LiDAR bundle adjustment} 



Bundle adjustment (BA) has been extensively studied in computer stereo vision and photogrammetry \cite{mur2015orb, qin2018vins}, where it involves refining the 3D scene geometry and the parameters of frame poses jointly, given a relatively coarse initial value. However, LiDAR-based BA is much less developed, and consequently, researchers in the robotics community have shown increasing interests in the problem of LiDAR bundle adjustment, aiming to ensure the global consistency of point clouds after pair-wise registration. Kaess \cite{kaess2015simultaneous} proposed using the plane-to-plane distance as the cost function, which requires obtaining accurate plane estimation from segmentation in each frame and is more suitable for dense point cloud from RGB-D cameras than for sparse LiDAR point clouds.

Point-to-plane distance is a more reasonable residual for sparse LiDAR point cloud. Ferrer \cite{ferrer2019eigen} proposes Eigen-Factors (EF), which optimizes the minimal eigenvalues of a homogeneous covariance matrix for the points on a plane, but has slow convergence due to applying first-order gradient descent for iterative optimization. Zhou \textit{et al.} \cite{zhou2020efficient} proposes planar bundle adjustment (PBA), 
{which requires to estimate the plane feature in each iteration by Schur complement similar to visual BA.} Huang \textit{et al.} \cite{huang2021bundle} modified the cost function by including an extra heuristic penalty term, which is not the true map consistency and require to extract accurate eigenvectors for the points corresponding to one pose on a plane feature. BALM \cite{liu2021balm} minimizes the eigenvalues in second derivative, converging much faster than EF, but has the drawback of requiring the enumeration each point in iteration. For this reason, BALM2 \cite{liu2022efficient} is proposed, which adopts point-to-plane as cost function, converges fast, eliminates features ahead, and has no singularity, adopting the concept of \textit{point cluster} to avoid the enumeration of each individual point. Thus, in this paper, we primarily adopt BALM2\footnote{\url{https://github.com/hku-mars/BALM}} as the LiDAR BA to compare different plane extraction methods.

\subsection{Plane extraction}

A common requirement in all BA methods reviewed above is the extraction of plane features from the input point clouds. Three main categories of plane extraction methods could be found in the litearture: Hough transform (HT), RANSAC, and region growing (RG).

\subsubsection{Hough transform}

Hough transform is a fundamental tool for identifying various shapes in 2D images \cite{hough1962method} and has been extended to 3D scenes \cite{borrmann20113d}. In order to extract the plane feature in 3D scene, HT first transforms the point $\mathbf p = (p_x, p_y, p_z)$ from Cartesian coordinates into Hough space $(\theta, \phi, \rho)$. The relationship is 
\begin{align}
	p_x \cdot \text{cos} \theta \text{sin}\phi + p_y \cdot \text{sin} \phi \text{sin} \theta + p_z \cdot \text{cos} \phi = \rho \label{eq:ht}
\end{align}
where $\theta \in [0, 2\pi]$ is the angle of normal vector in the $xy$-plane, $\phi \in [0, \pi]$ is the angle between $xy$-plane and the normal vector in $z$ direction, $\rho$ is the distance from origin to the plane. For the standard Hough transform (SHT), the Hough space is segmented into discrete cells. Each point is then transformed into Hough space, and the cells whose $(\theta, \phi, \rho)$ satisfy (\ref{eq:ht}) have their score increasing by one. Once all points finish voting, the cell with the highest scores are selected as the winning plane. 

However, due to the complexity depending on input points and the discretization of accumulators, high computation cost is a significant problem for SHT. Subsequent refined HT methods have been proposed to accelerate this process, including probabilistic HT (PHT), adaptive probabilistic HT (APHT), progressive probabilistic HT (PPHT) and randomized HT (RHT). Borrmann \textit{et al.} summarize these methods \cite{borrmann20113d} and their experiments indicate RHT is the best one. {Inspired by the kernel-based HT (KHT)} \cite{fernandes2008real} {which is designed for 2D images}, Limberger \textit{et al.} \cite{limberger2015real} present a deterministic technique that combines with octree and principal component analysis (PCA). The method achieves state-of-art real-time extraction with cost $O(n\text{log}n)$.

\subsubsection{RANSAC}

Random sample consensus (RANSAC) \cite{fischler1981random} is another popular model-based algorithm for iteratively extracting shapes. To detect planes, three points are randomly selected to compute a plane. The number of points which lie on the plane is counted to score the plane. This procedure is repeated iteratively until the probability of detecting the best plane exceeds the determined threshold. The plane with the highest score is then extracted as the final plane.

However, the randomness of choosing three points causes the standard RANSAC to suffer from a spurious-plane problem when noise and outliers exist. Schnabel \textit{et al.} \cite{schnabel2007efficient} refined RANSAC by using a localized sampling strategy and an optimized score function with normal estimation. Yang \textit{et al.} \cite{yang2010plane} integrated RANSAC and minimum description length (MDL) to prevent detecting wrong planes. Li \textit{et al.} \cite{li2017improved} improved RANSAC based on normal distribution transformation (NDT) cells to avoid the extraction of spurious planes.

\subsubsection{Region growing}

Region growing (RG) is a more efficient approach for plane extraction than Hough transform and RANSAC. The first step of region growing is to segment the point cloud into plenty of pieces and then pick a seed region to expand, by examining the co-planarity of its neighboring regions as growing rules. 

Poppinga \textit{et al.} \cite{poppinga2008fast} propose a region growing method that employs two {neighboring} points as the seed to grow. They extended the set by finding neighboring points at increasing distances from original region with a novel determination approach. However, treating single point as the growing unit is inefficient especially with the massive points. Thus, some authors proposed using voxel grids and octrees as region units to improve the efficiency and robustness. Deschaud \textit{et al.} \cite{deschaud2010fast} estimate the normal of each point and compute the score of local plane in each point. Finally, they select the best local seed plane as the seed region for voxel growing. Vo \textit{et al.} \cite{vo2015octree} use principal component analysis (PCA) to estimate each octree node as a plane or not. The ``planar cells" on the same plane are merged, and individual points of ``non-planar cells" are also added to the existing plane based on the distance of point-to-plane. Based on \cite{vo2015octree}, Araujo \cite{araujo2020robust} further presented a novel planarity test which is parameter adaptive and insensitive to outliers. 

The procedure of plane-based region growing \cite{vo2015octree, araujo2020robust} is similar to the plane extraction for LiDAR BA \cite{liu2021balm} and this work. However, the above region growing methods are not suitable for LiDAR BA due to reasons below:

\begin{enumerate}
\item Plane extraction for LiDAR BA is more time-sensitive, especially for local BA in LiDAR SLAM. Besides, the scene for LiDAR BA is always much larger than the experiments of above methods, which leads to higher requirements for efficiency. 
\item The segmented planes should not be too large. Large planes may increase the probability of {false plane association}, such as the large curved surfaces, which deteriorate the accuracy of localization.
\item The extracted points on each plane are not required to be too exhausted. Above RG approaches \cite{vo2015octree, araujo2020robust} have a procedure to extract points in the non-planar voxel, while it is unnecessary for LiDAR BA. On the other hand, it is always difficult to judge the boundary of planes as discussed in Section \ref{experiment1}. 
\end{enumerate}

\section{Methodology} \label{methodology}

This section mainly presents our methodology of plane extraction for LiDAR bundle adjustment. Firstly, we propose the novel plane determination approach. Next, the pipeline of our plane extraction is presented based on adaptive voxelization and merging.

\subsection{Plane determination}

In most LiDAR SLAM and plane extraction work, an efficient method to check co-planarity of a set of points is by utilizing principal component analysis (PCA) to obtain eigenvalues $(\lambda_1 \ge \lambda_2 \ge \lambda_3)$ from the points' covariance matrix. The ratio between the largest and smallest eigenvalue is compared to a threshold $\tau$. If the points lie on a plane, the following formula should be satisfied:

\begin{align}
	\lambda_1 \ge \lambda_2 \ge \lambda_3 \qquad
	\frac{\lambda_3}{\lambda_1} < \tau \label{pca}
\end{align}

This procedure is efficient regardless of the point cloud size, yet it has certain limitations. If the condition (\ref{pca}) is too strict, the recall rate is too low which is detrimental for the accuracy and robustness for BA optimization. Conversely, if the condition is relaxed, many non-planar points will be reserved, and the false-positive rate is raised. Generally, it is extremely difficult to choose an appropriate parameter, since condition (\ref{pca}) means the variance of thickness should be far less than that of length. It is the necessary but insufficient condition for plane determination. 

\begin{figure} [t]
	\centering
	\includegraphics[width=1\linewidth]{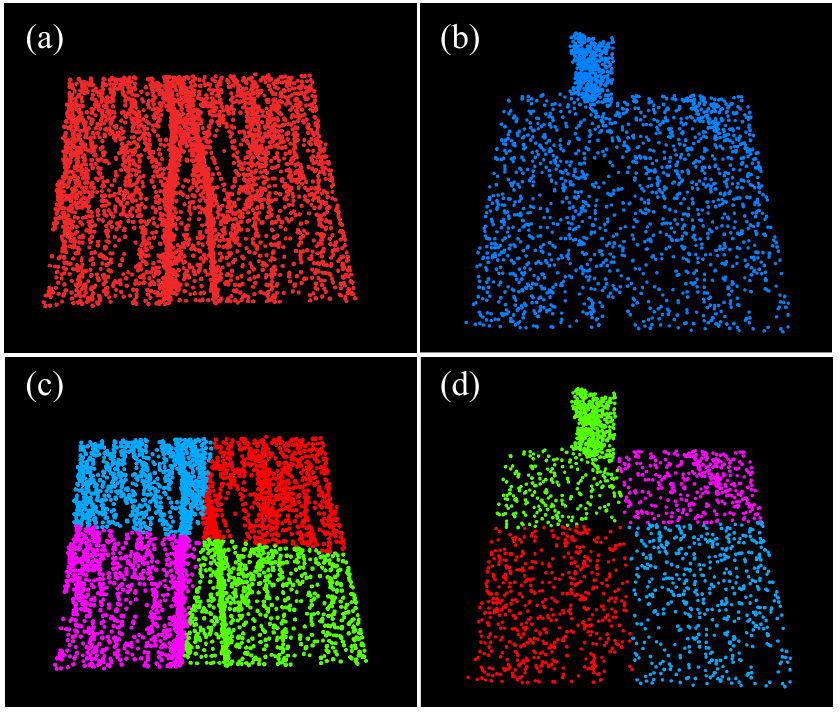}
    \caption{(a) True positive plane. (b) False positive plane if using the criterion in (\ref{pca}). (c) Smaller parts of a true plane are still planes. (d) Smaller parts of a false plane no longer form planes (i.e., the green part). }
    \label{fig positive}
\end{figure}

Therefore, we introduce an additional restrictive condition to determine the plane. In the process of observing LiDAR point cloud, we found that the false-positive point cloud, which satisfies condition (\ref{pca}), looks very similar to a plane but with obvious outlier points, as shown in Fig. \ref{fig positive}(b). In other words, for a true plane, if it is divided into smaller parts, each part should still be a plane. Inspired by this, we split the points set into quarters as evenly as possible. If each quarter has a similar ``thickness" to original points set, the set of points will be considered on the same plane. 

\begin{figure} [ht]
    \centering
    \includegraphics[width=1\linewidth]{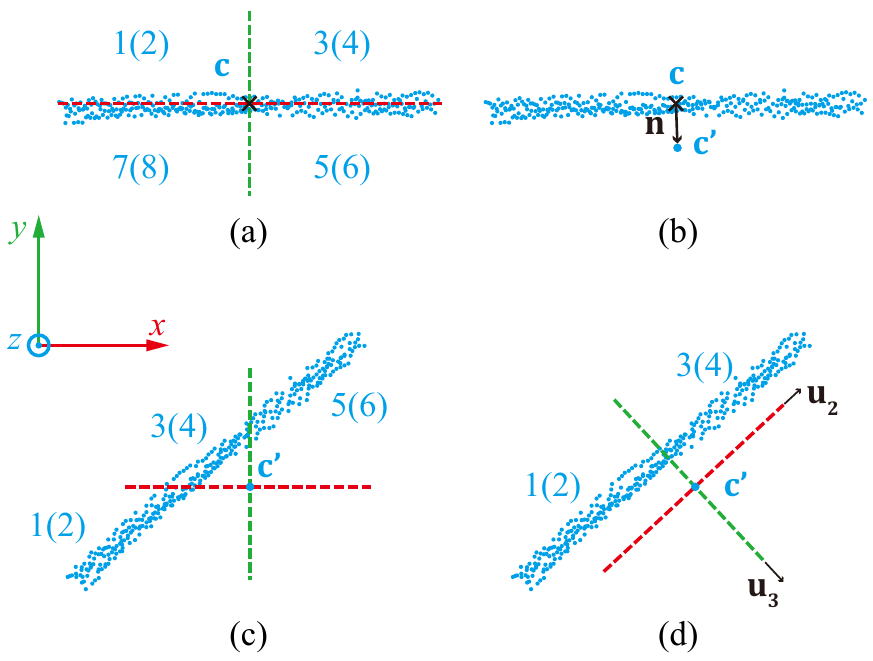}
    \caption{The side view of a set of points which are on the same plane. (a) Splitting on the centroid, which mistakenly separates the points into eight parts (only four parts are visible from the side view, while the other four un-visible parts are labelled in the parentheses.). (b) Translation of the split center from centroid along the normal in the distance of 5$\sigma$ ($\sigma = \sqrt{\lambda_3}$). (c) Splitting along $xyz$-axis, which separates the points into six parts (three parts can be seen from the side view). (d) Splitting along the plane normal, which separates the points into four parts evenly as expected (two parts can be seen from the side view).}
    \label{fig quarter}
\end{figure}

Due to the point noise and drift of poses, the points on one plane is seemingly like a ``cuboid" which has a obvious thickness as shown in Fig. \ref{fig quarter}. When splitting the points set at the centroid $\mathbf c$ along $xyz$-axis, we found that it may not always get the ideal split, as shown in Fig. \ref{fig quarter}(a). The reason is that the centroid is always inside the cuboid, and if plane normal is relatively parallel with axes which occurs frequently in the point clouds of ground and wall, there is a chance to mistakenly divide the point cloud belonging to one plane into two parts. 

To solve this issue, we move the centroid along the plane normal in the distance of $5\sigma$ thickness, where the plane normal $\mathbf n$ is the eigenvector corresponding to $\lambda_3$ and $\sigma = \sqrt{\lambda_3}$ in Fig. \ref{fig quarter}(b). In this way, the split center is well outside the ``cuboid", which avoids the split of a plane into two layers. 

{Another issue resides in the unnecessary more parts that could be split from the plane, which is shown in Fig.} \ref{fig quarter}(c). {Naively splitting the plane along the coordinate axes often separates the points into six uneven parts if the plane orientation is not well aligned with the coordinate axes. To address this issue, we note that the eigenvectors can well describe the plane orientation, splitting along the eigenvectors can always separate the points into four even parts as expected, as shown in Fig.} \ref{fig quarter}(d). 

Fig. \ref{fig positive}(c) and (d) are the split point cloud from Fig. \ref{fig positive}(a) and (b), in which (a) is true positive plane, and (b) is false positive plane that only satisfies formula (\ref{pca}). {The minimum eigenvalues of original points set can be seen the weighted average minimum eigenvalues of its four quarters. Hence, for the true plane, as shown in Fig. \ref{fig positive}(a) and (c), the original points and quarters have similar minimum eigenvalues and their ratio is approximately equal to $1$. However, for the false positive plane in Fig. \ref{fig positive}(b) and (d), the minimum eigenvalues of the original point set will be larger than the planar quarters and much smaller than non-planar quarters. Therefore, we can use the ratio of the minimum eigenvalue of the original points set and each its quarter as the criterion for plane determination.} According to this principle, the new plane determination condition is as follows:

\begin{align}
	\frac{\lambda_{3}}{\lambda_{1}} < \tau_1, \
	\frac{1}{\tau_2} < \frac{\lambda_{3}}{\lambda_{3l}'} < \tau_2; \ \tau_2 > 1, \forall l=1,2,3,4
	\label{new_judge}
\end{align}
where $\lambda_k$ is the $k$-th eigenvalue of original points set, $\lambda_{kl}'$ is the $k$-th eigenvalue of the $l$-th quarter, where $k$ is the index of eigenvalue $(k\in{1,2,3})$, $l$ is the index of quarters $(l\in{1,2, 3, 4})$. The first inequality in (\ref{new_judge}) is taken from (\ref{pca}) to ensure the original points set is in the shape of a plane. The second inequality in (\ref{new_judge}) guarantees that the difference of eigenvalue between the original points set and each quarter should not be too large.

\subsection{Adaptive voxelization and merging}
\begin{figure} [t]
	\centering
	\includegraphics[width=1\linewidth]{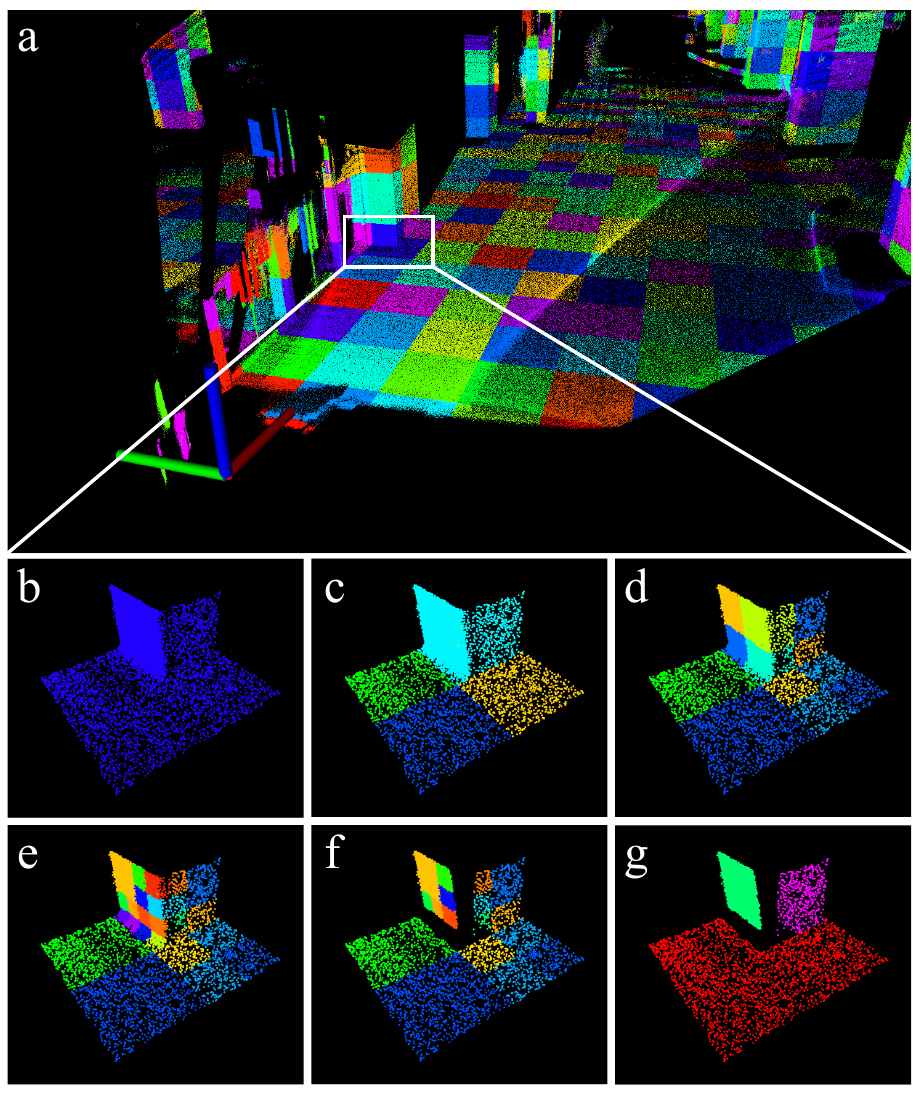}
	\caption{Adaptive voxelization and merging of planes. (a) shows the point cloud consisting of root voxel nodes. (b) is the root voxel which will be used to illustrate our theory. (c)-(e) is the first, second and third partition respectively. (f) removes the voxels which do not form qualified feature. (g) is the final merging result.}
	\label{fig merge}
\end{figure}

Here, we will give a brief introduction for adaptive voxelization proposed in \cite{liu2021balm}. The 3D space is firstly voxelized in a predetermined size where these initial voxels are called root voxels. To save memory and avoid processing empty voxels, hash map is adopted in which each non-empty voxel center corresponds to a hash key. For each root voxel, we apply formula (\ref{new_judge}) to determine whether the points are coplanar. If the points are coplanar, the voxel will be preserved; otherwise, the voxel will be divided into eight octants, and each octant is examined until the terminal condition (minimal points nubmer or octant size) is met. 

Adaptive voxelization is efficient and can always provide enough features to estimate states. However, it has one drawback: the partition of octree is irreversible and generates too many small planes. For example, the point cloud in the corner of wall consists of three planes perpendicular to each other. When partitioning the point cloud into multiple layers, the intersection parts of plane will be removed, but too many small planes are generated, as shown Fig. \ref{fig merge}(e). Theoretically, three planes are enough to describe the constraints and too many small planes will increase computational time.

To address this issue, we adopt plane merging method. After partitioning, we obtain a series of small planes in a root voxel, each contained in one leaf voxels of the octree. We enumerate all these planes in the same root voxels and merge them if they are on the same plane. Specifically, the first plane is added to the first group by default. We then estimate whether the second plane is coplanar with the plane in the first group. If it is, the second plane will be pushed into the first plane group; otherwise, a new group will be created. Similarly, the following planes are determined to be merged with an existing plane groups or create a new group. Fig. \ref{fig merge} illustrates the procedure of voxelization and merging. During adaptive voxelization, the centroid $\mathbf c_i$ and eigenvector $\mathbf u_{ik}$$(k=1,2,3)$ of a leaf voxel (hence its contained plane) are already computed. So these information can be re-used when determining if two planes are the same: if two planes, $\mathcal P_i$ and $\mathcal P_j$, are coplanar, the following condition should be satisfied,

\begin{align}
	\Big | \left\langle \mathbf u_{i3}, \mathbf u_{j3} \right\rangle & \Big |
	< \epsilon_1 \notag
	\\
	\Big | \left\langle \mathbf c_i - \mathbf c_j, \mathbf u_{i3}\right\rangle - \frac{\pi}{2} \Big | < \epsilon_2 &
	\quad
	\Big | \left\langle \mathbf c_i - \mathbf c_j, \mathbf u_{j3}\right\rangle - \frac{\pi}{2} \Big | < \epsilon_2 \label{two-plane-judge}
\end{align}
where $\epsilon_1$ and $\epsilon_2$ are two given parameters. The first inequality is to guarantee the parallel of two planes. The second inequality is to guarantee the distance of two parallel planes small enough.

\section{Experiments} \label{experiment}

\begin{figure} [t]
	\centering
	\includegraphics[width=1\linewidth]{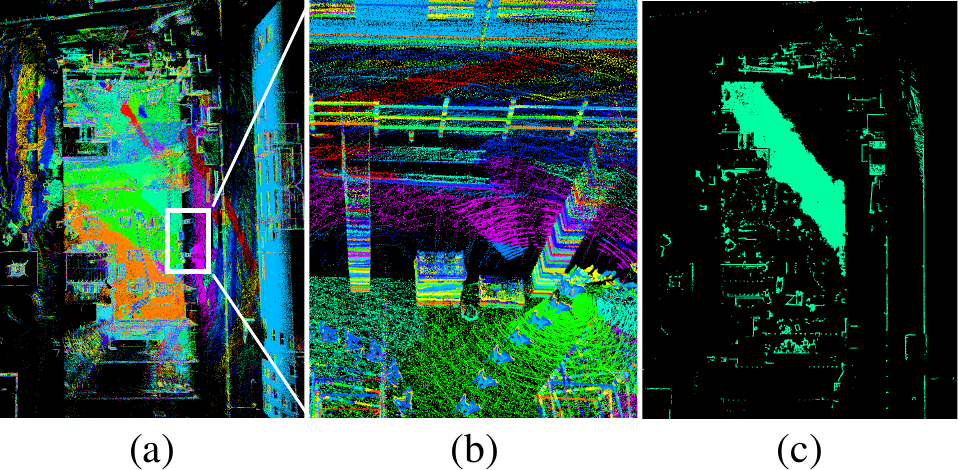}
	\caption{{(a) The plane extraction results of RANSAC in HILTI Exp01. (b) A local map of (a), which shows the point cloud is sliced into multiple layers parallel to ground. (c) A large plane extracted by RANSAC, which generates false plane association for LiDAR BA.}}
	\label{fig segment ransac}
\end{figure}

In this section, we present the experimental results to evaluate the efficiency and effectiveness of our proposed plane extraction approach. All the experiments were conducted on Ubuntu 20.04 with Intel 10750H CPU and 16 GB memory. We compared ours with four open-source approaches: random Hough transform (RHT) implemented by \cite{borrmann20113d}, RANSAC implemented by PCL\footnote{https://pointclouds.org/}, region growing (RG) implemented by \cite{araujo2020robust} and adaptive voxelization in BALM \cite{liu2021balm}. The point cloud is firstly voxelized in a prescribed size, even though it is not required for HT and RANSAC. The reasons for voxelization are as follows: 
{(1) For RHT, if applying it to the entire point cloud, it will take a very long time to process and even be stuck due to the too many points. (2) For RANSAC, as shwon in Fig. \ref{fig segment ransac}(b), one unexpected phenomenon is that it will slice the point clouds into multiple layers, which cannot adequately exploit the plane constraints on pillars. Since in this sequence the horizontal range is far larger than vertical range, RANSAC prefers to extract the planes parallel to horizontal direction, which includes more points than the vertical direction. Therefore, the extraction results are the planes parallel to horizontal direction by layer. The same reason also leads RANSAC to extract a large ground plane and falsely regards disjoint points as on the ground plane, as shown in Fig. \ref{fig segment ransac}(c). The regions are not connected and these points are not on the same plane actually, though the distance of point-to-plane is within the threshold ($0.03m$). These false plane detection will cause wrong plane associations, which lead the BA optimization to divergence. Therefore, we adopt the voxelization method for RHT and RANSAC.} The root voxel size is $1$m, minimum voxel size is $0.25$m and minimum points is 20. {RHT used the ``ball accumulator", with $\phi$ being discretized into $90$ bins, $\theta$ being discretized into $90$ bins and $\rho$ being discretized into $100$ bins ($N_{\phi}=90$, $N_{\theta}=90$, $N_{\rho}=100$).} The distance threshold of RANSAC is $0.03$m. For RG (Araujo), we use ``normal estimation" module in PCL to provide normals for each point, with $k$d-tree and searching radius $0.1$. Since the parameters in RG (Araujo) are automatically adjusted, we maintained the default parameters of the original paper \cite{araujo2020robust}. The parameters for our method are $\tau_1 = 0.0625$, $\tau_2 = 3$, $\epsilon_1 = 8^{\circ}$ and $\epsilon_2 = 10^{\circ}$.

We used the dataset HILTI 2022 \cite{zhang2022hilti} to evaluate precision and efficiency of these five methods. The accuracy of ground-truth in HILTI reaches millimeter scale, making it suitable for testing the performance of SLAM. The LiDAR used in this dataset is PandarXT-32. Moreover, we use FAST-LIO2 \cite{xu2022fast} to compensate the point cloud motion distortion and provide initial poses for plane extraction and BA optimization. 

\subsection{Extraction results} \label{experiment1}

\begin{figure*} [t]
	\centering
	\includegraphics[width=1\linewidth]{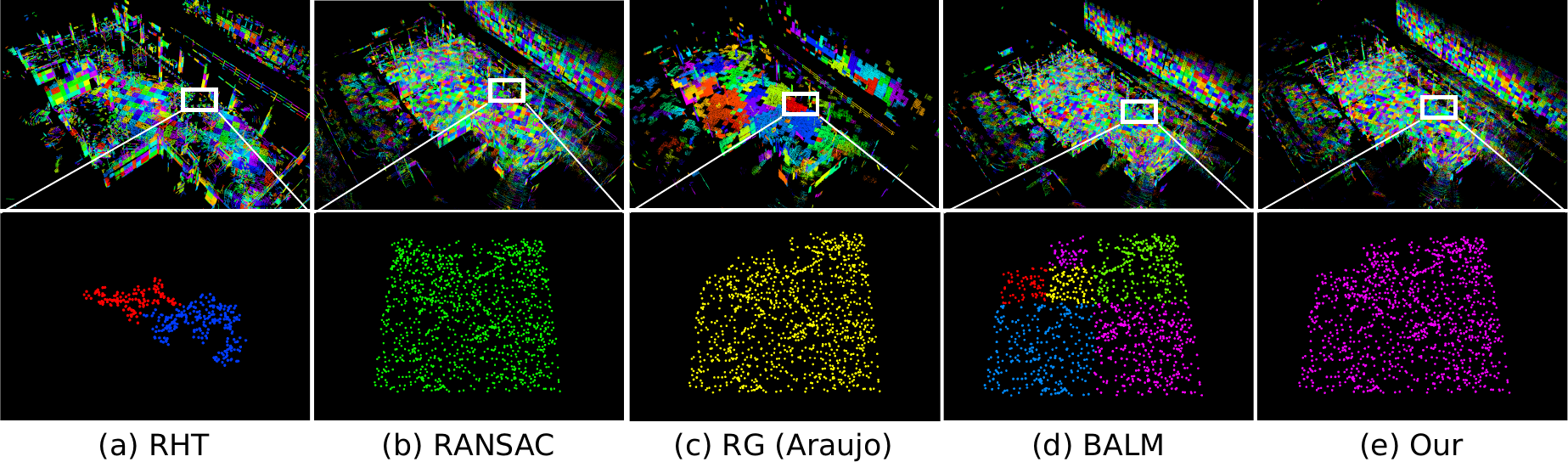}
	\caption{The plane extraction results of different methods in Exp01 of the HILTI 2022 dataset. The figure in first row is the overall extraction result. The figure in second row is a typical case of the extraction.}
	\label{fig case}
\end{figure*}

\begin{figure} [t]
	\centering
	\includegraphics[width=0.6\linewidth]{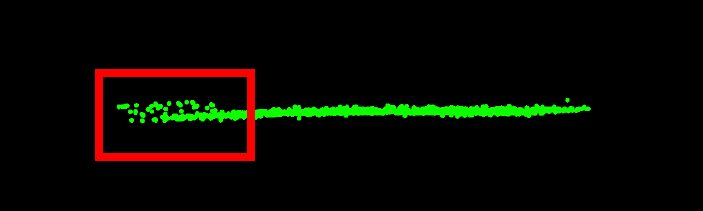}
	\caption{The side view of planar points in the lower figure of Fig. \ref{fig case}(b) }
	\label{fig ransac}
\end{figure}

It is usually difficult to obtain the ground-truth for planes contained in the LiDAR point cloud datasets. Further, the plane extraction mainly serves for LiDAR BA, rather than being a deliverable by itself. Therefore, we only conduct a qualitative analysis to compare the extraction results instead of quantitative comparison in other plane extraction works. The first row in Fig. \ref{fig case} is the overall extraction result in sequence Exp01, and the second row in Fig. \ref{fig case} presents a typical case in plane extraction. {The planes extracted by RG (Araujo) are a bit different from others with larger planes, due to its region growth.}

{As shown in the second row of Fig. \ref{fig case}(d), BALM \cite{liu2021balm} split the point cloud into multiple small pieces, which are obviously on the same plane. These small planes will increase the computing time of LiDAR bundle adjustment.} RHT and RG can remove these outlier points, but they also take away some points which is obviously on the plane, as shown in Fig. \ref{fig case}(a) and (c). The low recall rate is deteriorating the accuracy of LiDAR BA due to the lack of the constraints imposed by these points. RANSAC removes the outlier points that are beyond the certain distance (0.03m in this experiment). Likewise, our approach gets rid of the outlier part directly by adaptive voxel and merging.


Compared with our method, RANSAC extracts planes more exhaustively, which makes it suitable for applications such as 3D reconstruction that requires high recall to avoid cavity. However, for LiDAR BA, our approach is more suitable due to two reasons: (1) for accuracy: The target of plane extraction for LiDAR BA is to provide accurate point association. If an object lies on the ground, after the LiDAR scanning, it is difficult to determine whether a point at the joint is on the ground or on the object. As shown in Fig. \ref{fig ransac}, which is extracted by RANSAC and the side view of Fig. \ref{fig case}(b), some points extracted in the red frame are not on the ground which may cause false points association and decrease the final accuracy of LiDAR BA. (2) for efficiency: Removing the outlier part directly is more time saving. The logic is simpler and it does not require to traverse each point to determine whether on the plane. The time consuming results in Section \ref{experiment2} can also verify this point. 

\subsection{Accuracy evaluation for LiDAR BA} \label{experiment2}

Since the plane extraction is designed for LiDAR BA, we evaluate the plane extraction quality by the BA accuracy. Compared to other LiDAR methods, BALM2, which adopts point-to-plane as the cost function, has the advantage of converging fast, eliminating plane feature before the optimization, having no singularity and not requiring to enumerate each individual point. Thus, BALM2 is used here to evaluate these plane extraction methods. 

There are totally 16 sequences in HILTI 2022 dataset, but we selected {9} sequences to compare. For the remaining sequences, degeneration occurs in the narrow staircases, which requires extra IMU data for coverage and its treatment is beyond the scope of this paper. The results are presented in Table \ref{tb:results}.

\begin{table*} 
	\caption{Absolute trajectory error (ATE) and running time for different methods.}
	\label{tb:results}
	\centering
	{
		\begin{tabular}{clrrrrrrrrrr}
			\toprule
			& Methods &Exp01 & {Exp02} & Exp04 & Exp05 & Exp06 & Exp07 & Exp11 & Exp14 & Exp21 & Mean \\
			\midrule
			\multirow{5}{*}{ATE (cm)}
			& RHT    & 1.65 & 2.32 & 1.87 & 2.04 & 5.64 & 6.95 & 0.85 & 5.92 & 4.43 & 3.52\\
			& RANSAC & \textit{1.18} & 1.97 & 1.73 & 1.95 & \textbf{3.68} & 5.23 & 0.67 & 4.15 & 3.88 & 2.72\\
			& RG (Araujo) & 1.83 & 2.25 & 1.89 & 2.03 & 5.90 & 7.90 & 2.62 & 5.44 & 4.53 & 3.82 \\ 
			& BALM   & 1.50 & \textit{1.80} & \textit{1.62} & \textit{1.83} & 3.89 & \textit{4.20} & \textit{0.65} & \textit{3.83} & \textit{2.18} & \textit{2.39} \\
			& Ours   & \textbf{1.09} & \textbf{1.74} & \textbf{1.59} & \textbf{1.79} & \textit{3.79} & \textbf{4.05} & \textbf{0.59} & \textbf{3.52} & \textbf{1.71} & \textbf{2.21} \\ 
			\midrule
			\multirow{5}{*}{\shortstack{Times (s)\\(plane extraction)}} 
			& RHT    & 471.32 & 833.21 & 186.20 & 185.54 & 179.61 & 84.40 & 204.47 & 36.88 & 411.01 & 288.07 \\
			& RANSAC & 5.52 & 13.46 & 4.33 & 4.47 & 5.25 & 2.75 & 5.46 & 2.25 & 3.79 & 5.25 \\
			& RG (Araujo)    & 53.26 & 147.73 & 47.02 & 43.98 & 43.71 & 54.01 & 52.17 & 28.14 & 27.03 & 66.41 \\
			& BALM   & \textbf{3.31} & \textbf{7.60} & \textbf{1.62} & \textbf{1.48} & \textbf{1.69} & \textbf{0.60} & \textbf{1.56} & \textbf{0.27} & \textbf{2.14} & \textbf{2.92} \\
			& Ours   & \textit{3.86} & \textit{8.17} & \textit{1.82} & \textit{1.66} & \textit{1.91} & \textit{0.81} & \textit{1.78} & \textit{0.36} & \textit{2.62} & \textit{3.26} \\
			\midrule
			\multirow{5}{*}{\shortstack{Times (s)\\(total)}}
			& RHT    & 482.06 & 891.43 & 190.62 & 189.51 & 188.06 & 87.61 & 216.16 & 34.96 & 420.06 & 377.92 \\
			& RANSAC & 16.79 & 79.46 & 11.81 & 10.09 & 16.86 & 9.10 & 18.72 & 3.22 & 16.24 & 277.77 \\
			& RG (Araujo)   & 60.65 & 201.66 & 51.89 & 49.38 & 53.93 & 59.57 & 61.72 & 28.92 & 36.61 & 83.79 \\
			& BALM   & \textit{15.57} & \textit{72.88} & \textit{8.54} & \textit{9.35} & \textit{14.77} & \textit{7.53} & \textit{18.35} & \textit{1.44} & \textbf{14.81} & \textit{25.29} \\
			& Ours   & \textbf{15.11} & \textbf{70.26} & \textbf{8.44} & \textbf{8.92} & \textbf{14.40} & \textbf{7.45} & \textbf{16.05} & \textbf{1.36} & \textit{15.76} & \textbf{24.39} \\
			\bottomrule
		\end{tabular}
	}
\end{table*}

In this table, although RANSAC achieves better results on sequence Exp06, our method obtain the minimum absolute trajectory error (ATE) in most sequences, as well as on average. Our results are better than RHT, RG, and BALM on all the sequences steadily, owing to the high accuracy of our plane determination. As expected, BALM is faster than our method in terms of plane extraction, as we employ more steps for plane determination and merge the small planes in each root voxel. The reason why RG (Araujo) takes more time even than RANSAC is that the RG method in \cite{araujo2020robust} requires the normal vector of each point for plane determination. However, the raw points from LiDAR lack the normal information, necessitating the computation of the normal vector for each point using $k$d-tree, which is very time-consuming. When looking at the total time, which includes the time of plane extraction and LiDAR BA optimization, our method is faster than BALM in most sequences and on average, because by merging the small planes in each voxel, the number of planes is less than BALM, which will save time on BA optimization. Thus, our method achieves the best results in terms of accuracy and total time cost for LiDAR BA.

\section{Conclusion and Future Works} \label{conslusion}

This paper presented an efficient approach for plane extraction for LiDAR bundle adjustment. To improve the accuracy of plane determination, based on PCA we added extra restrictive conditions by splitting the candidate points into four quarters. The function of merging which can save the total time for LiDAR BA, was also appended with the adaptive voxelization. The experiments validated the efficiency and accuracy of the proposed approach. While the plane extraction is designed for LiDAR bundle adjustment, this system can also be applied on the LiDAR odometry (LO) or LiDAR inertial odometry (LIO) systems. In future work, we plan to investigate these applications of voxelized point cloud map for LiDAR (inertial) odometry, bundle adjustment and loop closure detection.

\bibliography{bare_jrnl}

\end{document}